\documentclass[10pt,journal,a4paper,final,oneside,twocolumn]{IEEEtran}
\usepackage{times}
\usepackage{graphicx}
\usepackage{cite}
\usepackage{color}
\usepackage{psfrag}
\usepackage{subfigure}
\usepackage{amssymb}
\usepackage{epsfig}
\usepackage{pifont}
\usepackage{amsmath}
\usepackage{array}
\usepackage[T1]{fontenc}
\usepackage{comment}
\usepackage{acronym}
\usepackage{multirow}
\usepackage{epsfig}
\usepackage{booktabs}

\usepackage{algorithm}
\usepackage{algorithmic}

\usepackage{amsthm,amsmath,amssymb}

\usepackage{mathrsfs}

\makeatletter
\newcommand{\removelatexerror}{\let\@latex@error\@gobble}
\makeatother

\begin{document}

\title{SemAI: Semantic Artificial Intelligence-enhanced DNA storage for Internet-of-Things}


\author{\IEEEauthorblockN{
                Wenfeng Wu, \emph{Student~Member, IEEE},
			Luping Xiang, \emph{Member, IEEE},
   			Qiang~Liu, \emph{Member, IEEE},
			and~Kun~Yang}, \emph{Fellow, IEEE}
			\vspace{-0.5 cm}\\

\thanks{This work was supported in part by the Natural Science Foundation of China under Grant 62301122 and Grant 62071101; in part by the Fundamental Research Funds for the Central Universities under Grant ZYGX2019J001; in part by the Sichuan Science and Technology Program under Grant 2023NSFSC1375. \textit{(Corresponding author: Luping Xiang.)}}
        \thanks{Wenfeng Wu and Luping Xiang are with the School of Information and Communication Engineering, University of Electronic Science and
        Technology of China, Chengdu 611731, China, email: wenfengwu@std.uestc.edu.cn, luping.xiang@uestc.edu.cn. }
        \thanks{Qiang Liu is with the Yangtze Delta Region Institute (Quzhou), University of Electronic Science and Technology of China, Quzhou, Zhejiang 324000, China, email: liuqiang@uestc.edu.cn.
        }

        \thanks{Kun Yang is with the School of Computer Science and Electronic Engineering, University of Essex, CO4 3SQ Colchester, U.K, and also with the School of Information and Communication Engineering, University of Electronic Science and Technology of China, Chengdu 611731, China, e-mail: kunyang@essex.ac.uk).}
	}

\maketitle

\begin{abstract}
  In the wake of the swift evolution of technologies such as the Internet of Things (IoT), the global data landscape undergoes an exponential surge, propelling DNA storage into the spotlight as a prospective medium for contemporary cloud storage applications. This paper introduces a Semantic Artificial Intelligence-enhanced DNA storage (SemAI-DNA) paradigm, distinguishing itself from prevalent deep learning-based methodologies through two key modifications: 1)  embedding a semantic extraction module at the encoding terminus, facilitating the meticulous encoding and storage of nuanced semantic information; 2) conceiving a forethoughtful multi-reads filtering model at the decoding terminus, leveraging the inherent multi-copy propensity of DNA molecules to bolster system fault tolerance, coupled with a strategically optimized decoder's architectural framework. Numerical results demonstrate the SemAI-DNA's efficacy, attaining  2.61 dB Peak Signal-to-Noise Ratio (PSNR)  gain and 0.13 improvement in Structural Similarity Index (SSIM)  over conventional deep learning-based approaches. 

\end{abstract}


\section{Introduction}

Internet of Things (IoT)   has permeated various domains, such as smart homes, industrial automation, and smart cities~\cite{20226GIot}. However, as the massive distributed sensors in IoT systems generate massive data that require efficient integration, storage, processing, and utilization, data management imposes a significant challenge in the implementation of IoT~\cite{2017Iotdata}. The key challenge lies in how to extract valuable information from the vast amount of data, enhancing the efficiency of data integration and storage.

Additionally, the data generated by IoT devices require a reliable and efficient storage system for storage and management. The significance of storage lies in providing data persistence and accessibility, enabling data to be queried, analyzed, and utilized. Currently, traditional storage media with cloud storage as a core application (such as magnetic storage, optical storage, solid-state storage, etc.) face technological bottlenecks in power consumption, size, reliability, and effective storage time~\cite{2022DNABased}. In recent years, DNA storage has garnered attention from many researchers due to its distinct characteristics~\cite{2024Tutorial}.

Specifically, compared with traditional storage media, DNA molecules exhibit significant advantages in data storage~\cite{2001Long,Extance2016HowDC}: (1) \textit{Ultra-high storage density.} The storage density of DNA molecules can reach $10^{19}  ~\mathrm{bit/cm}^3$, surpassing traditional storage media by six orders of magnitude. (2) \textit{Extremely long lifespan.} Data stored in DNA may persist for millennia without specific artificial intervention. (3) \textit{Remarkably low maintenance costs.} The footprint, resources, and energy required for DNA storage are significantly smaller than those for traditional storage media, resulting in minimal maintenance costs. Furthermore, the biochemical reactions and inherent operations of DNA molecules demonstrate substantial parallelism. Despite current drawbacks such as high read/write costs and slow read/write speeds, DNA is still considered the optimal choice for storing over $60\%$ of global cold data, making it a potential storage medium for current cloud storage applications\cite{2018article}.
Vast amounts of sensor data, images, and video streams are expected to be encoded and stored in tiny DNA molecules, achieving extremely high information density. 

{Currently, with the rapid development of DNA synthesis and sequencing technologies, DNA storage has witnessed numerous groundbreaking advancements}~\cite{2012Next,2013Towards,2015Robust,2017DNA,2017CRISPR,2018Random, Koch2019ADS}. In 2012, Church \textit{et al.}~\cite{2012Next} from Harvard Medical School first stored 0.65 MB of data in vitro, sparking a surge of research interest in DNA storage. Then in 2013, Goldman \textit{et al.}~\cite{2013Towards} increased the capacity of stored data to 739~\text{KB} of data. Grass \textit{et al.}~\cite{2015Robust} translated 83~\text{KB} of information to 4991 DNA segments and encapsulated in silica, which proved that data can be archived on DNA for millennia under a wide range of conditions.
In 2017, Shipman \textit{et al.}~\cite{2017CRISPR} encoded a movie message into living cells through CRISPR technology. 
In 2018, Microsoft and the University of Washington achieved random access to 200 MB of stored data~\cite{2018Random}. 
In 2019, Julian \textit{et al.}~\cite{Koch2019ADS} devised a "DNA-of-things" storage architecture to produce materials with immutable memory. They successfully stored and retrieved a 1.4~\text{MB} video in plexiglass spectacle lenses.

\begin{table*}
  \centering
  \caption{Contrasting the contributions of this work to the literature}
  \begin{tabular}{l |c |c| c| c |c|c|c|c|c}
    \toprule
    Contributions                & This work & \cite{2024Tutorial} &\cite{2017DNA}& \cite{nguyen2021capacity}& \cite{qin2024robust}& \cite{2023Deep} 
 & \cite{siddaramappa2019dna} & \cite{2020Algorithmic} &\cite{2021Nanonetworks}\\
    \midrule
    Large-AI model               & \checkmark &   &              &                      &                         &    & & &\\
     Multi-Reads-based correction  & \checkmark &    &             & &                           \checkmark&  &&&\\
    Semantic-based system               & \checkmark &   &              &                      &                     &        &&&\\
    Autoencoder-based system        & \checkmark & &   &                      &                           &  \checkmark&&&\\
    Biological constrain                & \checkmark &  & \checkmark      &                      \checkmark& &  \checkmark && \checkmark&\checkmark\\
    Complete coding scheme              & \checkmark & \checkmark & \checkmark      & \checkmark           & \checkmark                 & \checkmark & \checkmark& \checkmark&\checkmark\\
    \bottomrule
  \end{tabular}
  \label{tab.contributions}
\end{table*}

\begin{figure*}[!htbp]
  \centering
    \includegraphics[width=1\textwidth]{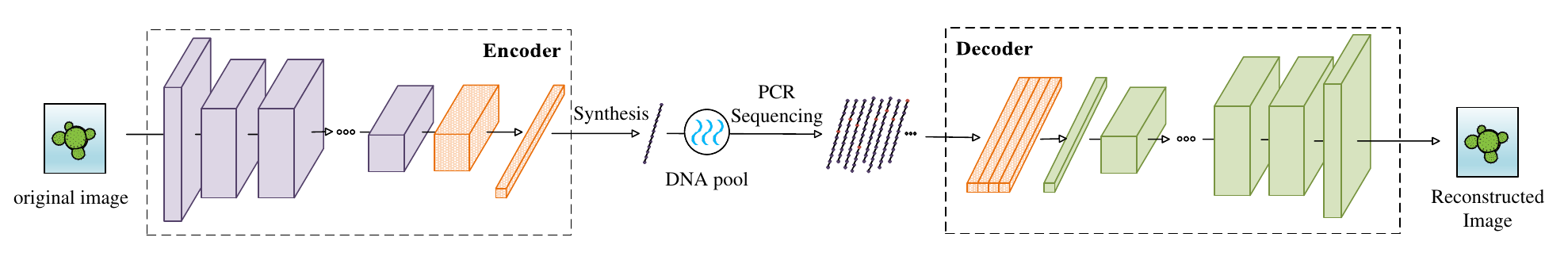}
    \caption{The system model of DJSCC-DNA~\cite{2023Deep}.}
\label{fig.model0}
\end{figure*}

DNA encoding is a crucial way of achieving DNA storage, wherein raw data is encoded into a sequence composed of the four bases ($\text{A, C, G, T}$) through specific mapping rules and can be reconstructed back to the original data\cite{2020research}. {The challenge in DNA encoding lies in designing high-density and fault-tolerant encoding strategies.} 
A common approach is to apply classical coding theories, such as utilizing Huffman coding and fountain codes for data compression~\cite{2013Towards,2017DNA,2019Author}, and incorporating Forward Error Correction (FEC) codes, such as Reed-Solomon (RS) codes, Low Density Parity Check (LDPC) codes, and Bose-Chaudhuri-Hocquenghem codes (BCH) codes to enhance fault tolerance~\cite{2015trends,2015Robust,2018Random,2019Author,2021An}. 
However, due to the distinct physical and chemical properties as well as constraints inherent in DNA sequences compared to conventional digital communication environments, such as limitations on sequence length, distribution ratios of base pairs, and length constraints on homopolymers, which pose additional challenges to conventional coding schemes~\cite{2021MINIMUN, nguyen2021capacity}. 
Moreover, noise forms introduced during the synthesis and sequencing processes of DNA sequences, such as base pair insertions and deletions~\cite{shomorony2021dna}, differ from those in traditional digital communication, rendering traditional coding schemes ineffective in rapidly and effectively identifying and correcting a large number of base errors. 
Therefore, some scholars have introduced deep learning (DL)  techniques into DNA encoding to address the complex characteristics and encoding requirements of DNA sequences, leveraging the powerful feature learning capabilities to improve encoding efficiency and accuracy~\cite{Generative, qin2024robust,2023Deep}.
In~\cite{qin2024robust}, a robust multi-reads reconstruction neural network was proposed to handle noise in DNA storage, including base substitutions, insertions, deletions, and chain breakage. In\cite{2023Deep}, a deep joint source-channel coding system for DNA storage, called DJSCC-DNA, was designed. As shown in Fig.~\ref{fig.model0}, DJSCC-DNA replaces traditional encoding and decoding with convolutional neural networks (CNN) and uses multiple copies of erroneous DNA chains for information retransmission at the decoding end, achieving high-density and fault-tolerant data storage. The system also incorporates an end-to-end optimization function to ensure DNA sequences adhere to biological constraints, specifically limiting GC content and homopolymer length.
Up to the present, research on utilizing DNA molecules for data storage primarily focused on syntactic data storage, aiming to ensure the complete preservation of data information. 

However, with the widespread deployment of IoT devices, the explosive growth of multimedia applications such as extended reality, autonomous driving, and intelligent surveillance has emerged. This rapid expansion poses tremendous challenges to the utilization of limited spectrum resources, prompting a shift towards semantic communication paradigms~\cite{2024joint}. Semantic communication differs from the traditional Shannon paradigm by focusing on conveying the actual meaning intended by the sender or information conducive to achieving common goals, whereas the Shannon paradigm is more concerned with the accurate reception of bits~\cite{strinati20216g}. With the rapid development of artificial intelligence (AI), the integration of knowledge bases and DL algorithms has found extensive applications in constructing models for semantic learning and communication~\cite{2021Semantic,zhang2023drl,weng2023deep,zhang2023predictive}. For example, Jiang~\cite{jiang2023large} proposed a semantic framework based on large-scale AI models, incorporating a knowledge base built on the Segment Anything Model (SAM) and introducing Attention-based Semantic Integration (ASI) to integrate semantic-aware data sources. 
In parallel, the current DNA image storage paradigm still emphasizes the encoding and restoration of the overall image, balancing high compression rates with the accuracy of image extraction. Inspired by this, to adapt to the continuously evolving IoT deployment environment, this paper shifts focus from storing comprehensive data information in DNA to prioritizing the storage of semantic information. By leveraging semantic communication principles, we aim to enhance the efficiency and effectiveness of data storage in IoT systems.

In the context of IoT, the need for efficient and scalable data storage solutions is more pressing than ever. This paper proposes a Semantic AI-enhanced DNA Storage (SemAI-DNA) scheme that utilizes DNA storage technology to encode the semantic information of data into DNA sequences, thereby making it possible to store or transmit large-scale data in the upcoming IoT era.
The key technical contributions of our investigation are contrasted to the literature in Tab.~\ref{tab.contributions} and are summarized as follows:

\begin{itemize}
\item We propose a large-scale AI model for DNA storage by utilizing the novel application of SAM and ASI techniques, which enables the extraction and storage of rich semantic targets from images. By leveraging the image semantic extraction module, the system efficiently achieves DNA image semantic storage, ensuring the accurate transfer of meaningful information.

\item By exploiting the unique characteristics of DNA multi-reads, we propose a multi-reads-based DNA chain screening scheme. To ensure reliability, the system evaluates and filters the multi-reads through isomorphic matching and $K$-mers comparison techniques, selecting highly trustworthy multi-reads for further processing. The error correction module utilizes parallel convolutional layers to extract multi-layered semantic information from DNA multi-reads, which are then fed into the decoder to reconstruct the semantic image.

\item Our numerical results demonstrate that the proposed SemAI-DNA scheme effectively achieves high-quality resolution of semantic information in images and high fault tolerance while adhering to biological constraints. Specifically, at a compression ratio of $R=0.25/3$ and the same base error rate $\gamma$ of $0.01$, our proposed scheme outperforms the existing DJSCC-DNA scheme~\cite{2023Deep}, which is utilized for storing the entire content of the image. SemAI-DNA achieves a $2.61~\text{dB}$ higher peak signal-to-noise ratio (PSNR)  and a higher structural similarity index (SSIM) of $0.13$. Furthermore, at a base pixel ratio $R=0.25/3$, our proposed SemAI-DNA scheme can withstand nearly four times the base error rate compared to DJSCC-DNA. 
\end{itemize}

\begin{figure*}[!htbp]
  \centering
  \includegraphics[width=1\textwidth]{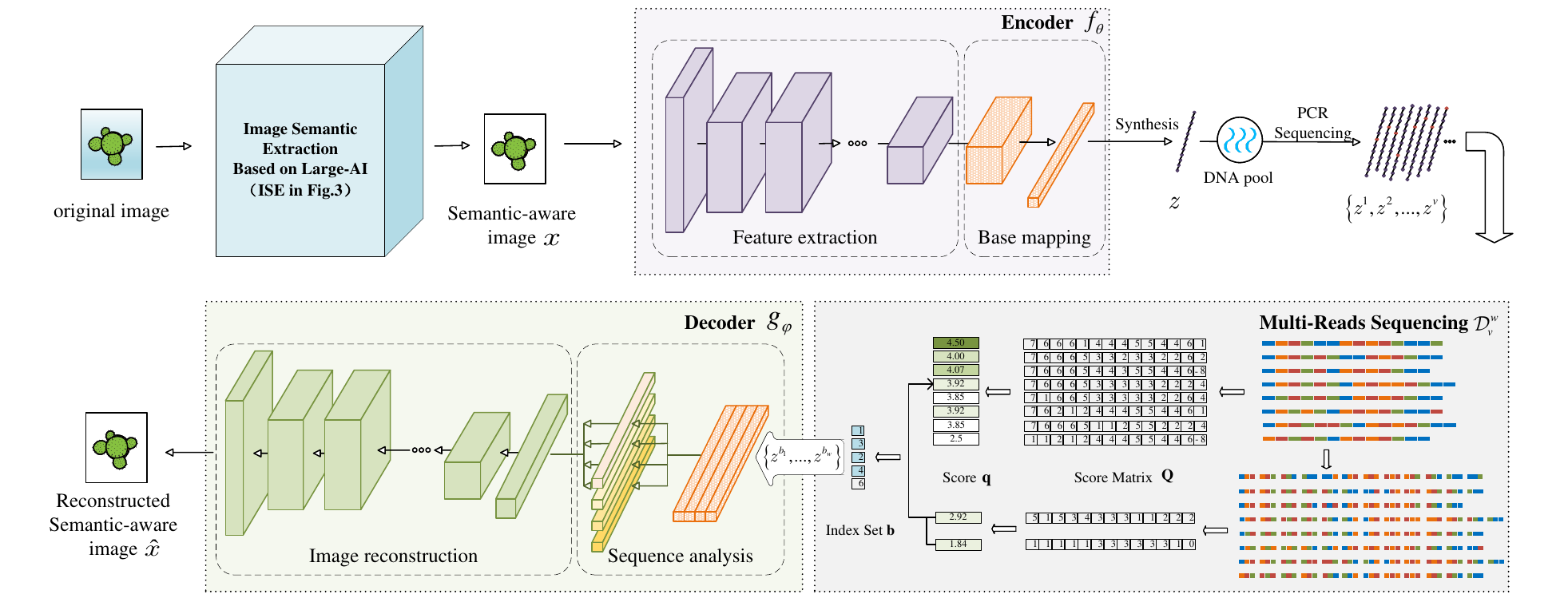}
  \caption{The proposed SemAI-DNA model.}
  \label{fig.model1}
\end{figure*}

\begin{figure}[!htbp]
  \centering
    \includegraphics[width=0.5\textwidth]{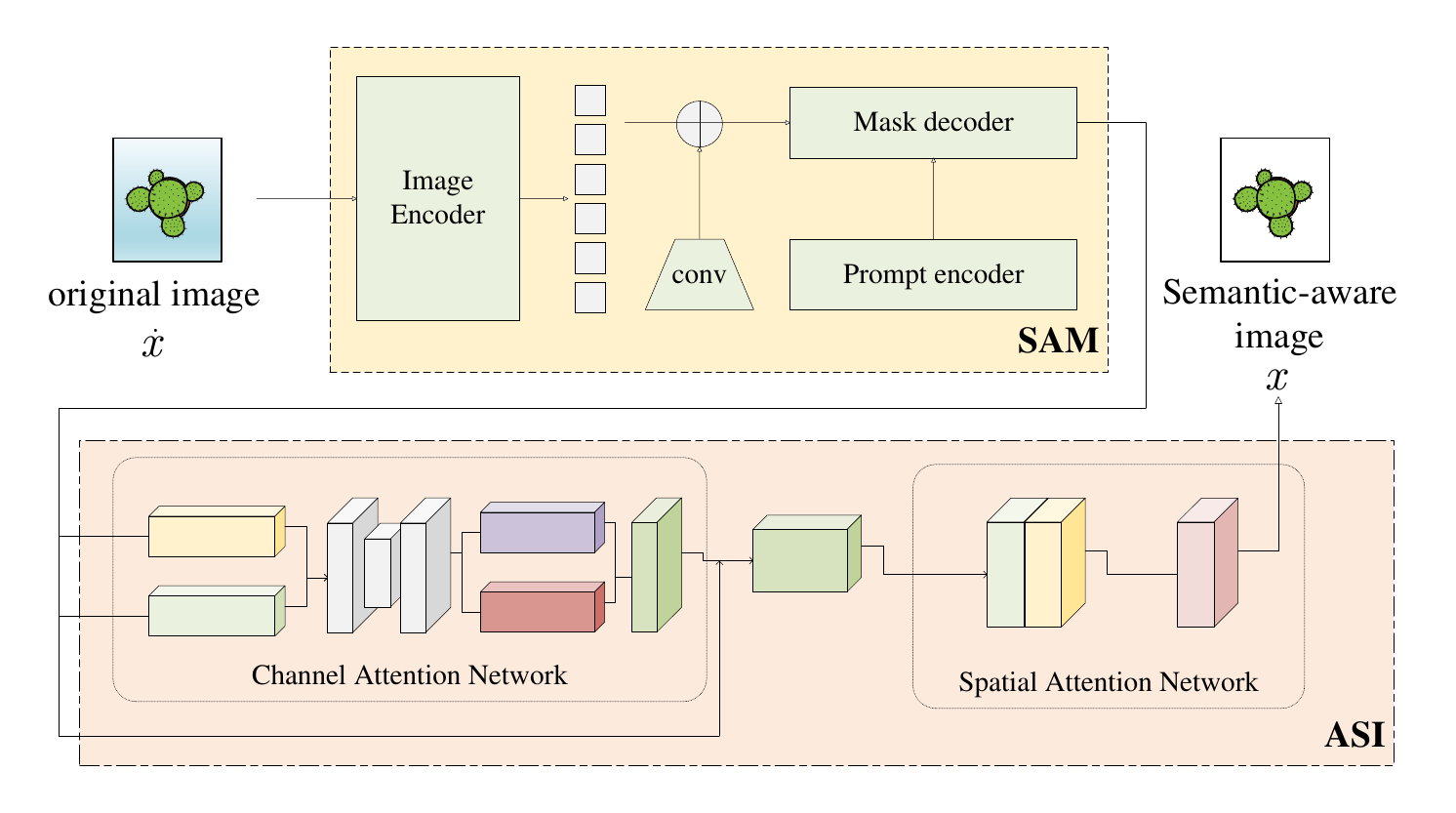}
    \caption{The structure of the ISE component.}
\label{fig.kbs}
\end{figure}

The structure of the remainder of this paper is as follows: 
Section II introduces the system model. Section III provides details on the DNA storage based end-to-end design. Numerical results are presented in Section IV, and the paper concludes in Section V.

\textit{Notations}: The DNA sequence is denoted by the letter $z$, e.g., $\pmb{z}$, $\hat{\pmb{z}}$. The $i$-th element of sequence $\pmb{z}$ is denoted as $z_i$. Multy-Reads of DNA are represented in the form $\{\pmb{z}^1, \pmb{z}^2, \pmb{z}^v\}$, where the superscript indicates the index of the specific sequence. The matrices and vectors are denoted by bold letters, e.g., $\mathbf{Q}$, $\mathbf{q}$. The elsement in the $i$-th row and the $j$-th column of matrix $\mathbf{Q}$ is written as $Q_{i,j}$; The $i$-th row of matrix $\mathbf{Q}$ is written as $\mathbf{Q}_i$; The $i$-th element of vector $\mathbf{q}$ is written as $q_i$.

\section{System Model}
\label{sec.System Model}


In this section, we introduce SemAI-DNA,  emphasizing the encoding and storage of semantic targets in images. On the one hand, this approach utilizes high-performance large AI models to achieve the extraction of semantic targets, eliminating redundant information outside the targets, thereby significantly reducing the storage data volume and improving storage efficiency. On the other hand, the multi-reads characteristics of DNA storage are leveraged to enhance decoding fault tolerance, ensuring high-quality storage of important semantic information under high compression rates. Finally, a potential IoT application scenario is proposed.

Building upon the research by~\cite{2023Deep}, this paper proposes a coding scheme based on DL for DNA image semantic storage. 
The proposed system model, as illustrated in Fig.~\ref{fig.model1}, primarily comprises modules such as the image semantic extraction  (ISE) based on the Large-AI module, the deep joint semantic-channel encoding and decoding module (Encoder-Decoder), and the Multi-Reads Screening  (MRS) module.
Firstly, the ISE utilizes a semantic recognition and integration model based on a shared semantic-aware knowledge base to extract the most meaningful semantic targets from the image.
Secondly, the encoder constructs a CNN
to convert the integrated semantic-aware image into DNA sequences. During the processes of DNA synthesis, Polymerase Chain Reaction (PCR) amplification, and sequencing, multiple DNA sequences are prone to various types of base errors, which significantly impact subsequent data recovery efforts. The MRS screens out more suitable sequences from multiple DNA strands for decoding. Lastly, the decoder analyzes the DNA sequences, mitigating the effects introduced by the DNA storage channel, and reconstructs the semantic-aware image.


\subsection{Image Semantic Extraction}
\label{sec.extraction}

The ISE module utilizes attention mechanisms and large-scale model to accurately capture the most crucial and representative semantic-aware image $\pmb{x}$ from the original image $\pmb{\dot{x}}$. As shown as Fig.~\ref{fig.kbs}, it primarily comprises two components~\cite{Jiang2023LargeAM}: 

$\emph{1)}$ The SAM module. We utilize SAM as a knowledge repository to achieve semantic segmentation of images. SAM~\cite{kirillov2023segment} is an indicative model, leveraging the largest segmentation dataset to date, the Segment Anything 1-Billion mask dataset, which comprises over 11 million images totaling more than one billion mask images. The model is designed as an interactive prompting model during training, enabling zero-shot learning transfer to new image distributions and tasks. The SAM architecture primarily consists of three components: an image encoder, a prompt encoder, and a mask decoder. It is capable of producing an effective segmentation mask given any segmentation prompt.

$\emph{2)}$ The ASI module, is an attention-based semantic integration module that emulates human perception by selectively choosing the most salient semantic segments, ultimately synthesizing a novel semantic perceptual image~\cite{jiang2023large}. It consists of channel attention network and a spatial attention network, which introduce attention mechanisms to identify important objects in the image of formaldehyde.

The  process can be expressed as follows:
\begin{align}\label{eq.ise}
  \pmb{x}=\mathcal{W}(\pmb{\dot{x}}),
\end{align}
where $\mathcal{W}$ presents the function of ISE.

\begin{figure*}[htbp]
  \centering
  \includegraphics[width=1\textwidth]{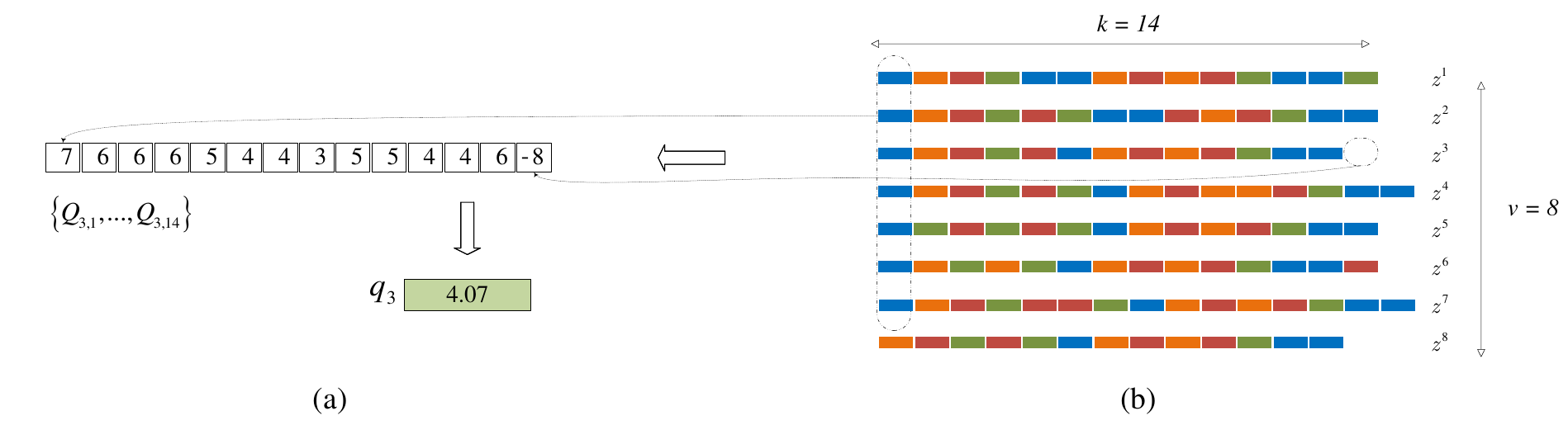}
  \caption{An example of calculating $\mathbf{Q}$ and $\mathbf{q}$, where $k=14,v=8$.}
\label{fig.screen}
\end{figure*}

\subsection{Encoder}
\label{sec.encoder}

The encoder encompasses two stages: feature extraction and base mapping. Initially, the encoder employs cascading convolutional layers to extract the latent features from the input image. By stacking multiple convolutional layers, the network acquires a more advanced level of feature representation. This hierarchical framework empowers the encoder to capture increasingly abstract and intricate semantic information.
Subsequently, the extracted latent features are meticulously mapped to their corresponding DNA sequences. The bases $\{\text{A, C, G, T} \}$ are represented by set $\{ 0, 1, 2, 3 \}$. This mapping procedure engenders a DNA sequence that encodes the semantic information of the input image by designating specific bases for the extracted features. 

Assuming the input image semantic sequence as $\pmb{x}\in\mathbb{R}^{n}$, where $n=H(\text{height})\times W(\text{width})\times C(\text{channels})$ is the number of pixels, the encoding process can be expressed as 
\begin{align}\label{eq.enc}
  \pmb{z}=f_{\theta}(\pmb{x}),
\end{align}
where $f_\theta:\mathbb{R}^{n} \xrightarrow{} \mathbb{Z}^{k}$ is the encoding function and $\theta$ denotes the parameter set of the encoder, and $k$ represents the base sequence size. The output sequence $\pmb{z} \in \mathbb{Z}^k$ will be generated by DNA synthesis to generate DNA strands for storage or transmission.\\

\subsection{Channel}

Most DNA synthesis techniques involve adding nucleotides in a specified order, and various errors may occur during this process~\cite{2020HEDGES,2021Fundamental}. For example, nucleotides may not be added to the designated location (base deletion); nucleotides may be added to the wrong location (base insertion); non-designated nucleotides may be added (base substitution). To facilitate data retrieval, primers are attached to both ends of the DNA chain. When the synthesis process is completed, the resulting DNA strands are then placed in an aqueous solution to generate a DNA pool. When reading the data, a specific DNA chain needs to be extracted from the main DNA pool, and PCR technology is used to amplify DNA chains containing the corresponding primers in the pool, thereby forming a new DNA library that includes multiple copies of the target DNA chain and a small amount of unrelated DNA chains. Target information can be obtained from this new library by sequencing. The main error type during sequencing is base substitution, while the probability of base insertion and deletion errors is significantly reduced.

In proposed SemAI-DNA model, we model the DNA synthesis, storage, amplification, and sequencing processes as a DNA channel, and simulate errors such as base substitution, insertion, and deletion. According to previous research~\cite{2020HEDGES, 2023Deep}, the probabilities of substitution, insertion, and deletion can be set to $17\%$, $40\%$, and $43\%$, respectively, and the total base error rate is denoted as $\gamma$. 
\begin{align}\label{eq.chl}
  {\hat{\pmb{z}}}=\mathcal{F}_\gamma(\pmb{z}),
\end{align}
where $\mathcal{F}_\gamma$ denotes the channel process with base error rate $\gamma$.
This process allows us to obtain a collection of multiple reading sequences that simulate the DNA channel and contain the specified information $\pmb{z}$, denoted as $\{\hat{\pmb{z}}^1,\hat{\pmb{z}}^2,..., \hat{\pmb{z}}^v\}$, where $v$ is the number of multiple reading sequences.
\\
\subsection{Multi-Reads Screening}
\label{sec.Screening}

The MRS module selects $w$ sequences $\{\hat{\pmb{z}}^{b_1},\hat{\pmb{z}}^{b_2},..., \hat{\pmb{z}}^{b_w}\}$ from $v$ sequences and converts them back to the digital sequence group as input for the decoder, which can be expressed as
\begin{align}\label{eq.chl2}
  \{\hat{\pmb{z}}^{b_1},\hat{\pmb{z}}^{b_2},..., \hat{\pmb{z}}^{b_w}\}=\mathcal{D}_v^w(\{\hat{\pmb{z}}^1,\hat{\pmb{z}}^2,..., \hat{\pmb{z}}^v\}).
\end{align}
where $\mathbf{b}$ is the vector of the index of the selected sequences and $\mathcal{D}_v^w$ denoted the process of selecting $w$ sequences from $v$ sequences.

Since DNA sequences have the unique characteristic of multi-copy generation during extraction, we propose to select high-reliability sequences from multiple sequence groups, rather than merely increase the number of sequences from multiple readings and not prioritize sequence quality in the selection process for data restoration, as it did in existing literture~\cite{2023Deep}. 
In this paper, sequences with fewer insertion, deletion, or substitution errors in multiple reads are considered highly reliable and more conducive to neural network decoding.
Preliminary screening of sequence groups reduces the likelihood of including highly contaminated sequences, thereby enhancing the accuracy and integrity of the restored data. 
The proposed MRS method is particularly well-suited for multi-read scenarios where the channel conditions are similar or identical.

The core idea of the scheme is to evaluate the reliability of sequences by comparing them and selecting sequences with lower levels of noise contamination for decoding. The specific operations are as follows:

$\emph{1)}$ Initialize a score matrix $\mathbf{Q}$ of size $v \times k$. The received sequence group is denoted as $\{\hat{\pmb{z}}^1,\hat{\pmb{z}}^2,..., \hat{\pmb{z}}^v\}$, where the number of sequences is $v$, and the length of the sequences is approximately $k$. The inconsistent lengths of the sequences are due to variations in the number of base insertions and deletions. 

$\emph{2)}$ Compute the values of $\mathbf{Q}$ based on the distribution of bases at the same positions among sequences. The number of occurrences of the $j\text{-th}$ base in the same column was counted and assigned to ${Q}_{i,j}$. If the base is missing at that position, assign a penalty value of $-v$ to ${Q}_{i,j}$.

$\emph{3)}$ Normalize the values of Q and calculate the row sums to obtain a score vector $\mathbf{q}$ for the $v$ sequences. 

$\emph{4)}$ Sort the elements of $\mathbf{q}$ in descending order to obtain the index set  $\mathbf{b}$ of the top $w$ scoring sequences, where $\mathbf{b} \in \{1, 2, 3, ..., v\}$.

$\emph{5)}$ Convert the sequences corresponding to the top $w$ indices in $\mathbf{b}$ into numerical sequence groups according to the {$\{\text{A, C, G, T}\} \Rightarrow \{1,2,3,4\}$ conversion rule, which will serve as the input for the decoder. If a sequence's length is less than $k$, pad it with zeros; if it exceeds $k$, ignore the excess part.

  \begin{algorithm}[H]
      \caption{Multi-Reads Screening}
      \begin{algorithmic}[1]
        \REQUIRE Initial $\{\hat{\pmb{z}}^1,\hat{\pmb{z}}^2,..., \hat{\pmb{z}}^v\}$ as a char matrix $\mathbf{X}_{v \times k}$
        \ENSURE sequence groups $\{\hat{\pmb{z}}^{f{b}_1},\hat{\pmb{z}}^{{b}_2},...,\hat{\pmb{z}}^{{b}_w}\}$
        \STATE Initial $\mathbf{Q} \leftarrow -v$  
        \FOR{$i=1$ to $v$}
          \FOR{$j=1$ to $k$}    
              \STATE ${Q}_{i,j} \leftarrow $  the number of char ${X}_{i,j}$ in the vector $\mathbf{X}_j$
          \ENDFOR
          \STATE ${q}_i \leftarrow $ the mean of $\mathbf{Q}_{i}$
        \ENDFOR
        \STATE $\mathbf{b}  \leftarrow $  sorts $\mathbf{q}$ and returns the element index sequence
        \IF{${q}_{{b}_w} = {q}_{{b}_{w+1}}$} 
        \STATE Indices to be rescreened: $\mathbf{p} \leftarrow $ Index of $\mathbf{q}_{{b}_w}$ in $\mathbf{q}$
        \STATE Number to be rescreened: $t \leftarrow$  size of $\mathbf{p}$
        \STATE Do $K$-mers, initial as a character matrix $\mathbf{X}'_{v \times (k-K+1)}$
        \STATE Initial $\mathbf{Q}' \leftarrow -v$
        \FOR{$i=\{\mathbf{p}_1,...,\mathbf{p}_t\}$}
          \FOR{$j=1$ to $k-K$}
            \STATE ${Q}'_{i,j} \leftarrow $  the number of ${X}'_{i,j}$ in the vector $\mathbf{X}'_j$
          \ENDFOR
          \STATE ${q}'_i \leftarrow $ the mean of $\mathbf{Q}'_{i}$
        \ENDFOR
        \STATE update $\mathbf{b}$ according to $\mathbf{q}'$
        \ENDIF
        \STATE Retains only the first $w$ elements of $\mathbf{b}$.
      \end{algorithmic}
  \end{algorithm}

Fig.~\ref{fig.screen} gives an illustrative example of the score calculation process. Specifically, on the right-hand side, there are 8 sequences, each denoted by a different color representing the four bases. The lengths of the sequences are approximately 14. The left-hand side of the figure displays the computation results of $\{{Q}_{3,1},...,{Q}_{3,14}\}$ in the matrix, while the computation result of ${q}_3$ is shown below. The complete results of $\mathbf{Q}$ and $\mathbf{q}$ can be found in Fig.~\ref{fig.model1}, where ${q}_4$ and ${q}_6$ have the same score but different sequences. If it is necessary to select the top four high-scoring sequences for decoding, and there are five sequences that have scored well, it becomes difficult to prioritize between ${q}_4$ and ${q}_6$. In such a situation, it is necessary to reevaluate these two sequences that have the same score.

When there are multiple elements in the $\mathbf{q}$  score vector that have the same $w$-th largest value, leading to the size of the index set $\mathbf{b}$ being greater than $w$, it is necessary to make decisions regarding which indices will be retained for filtering. Although these groups of sequences have the same score, their base arrangements may differ, and their reliability can be evaluated based on the specific distribution of bases. The specific method is as follows:

$\emph{1)}$ Assuming there are $t$ sequences that need to be decided, 
all sequences are divided into polymers of length K (a process referred to as $K$-mers, following the concept described in \cite{chor2009genomic}). This involves dividing the reads into subsequences of length $K$, where the first base of the next subsequence is the second base of the previous subsequence. Fig.~\ref{fig.kmer} illustrates the $K$-mers process, where $K=3$.

$\emph{2)}$ Compare these subsequences using a similar method as sequence comparison, and independently evaluate the score matrix $\mathbf{Q}'$ and score vector $\mathbf{q}'$  for these $t$ sequences.

$\emph{3)}$ The elements with high scores in $\mathbf{q}'$ will be retained in $\mathbf{b}$.

The complete process of obtaining $\mathbf{b}$ is illustrated in Fig.~\ref{fig.model1}. Due to the same score of ${q}_4$ and ${q}_6$, when selecting 4 sequences for decoding, a second evaluation was performed on $\hat{\pmb{z}}^4$ and $\hat{\pmb{z}}^6$, and ultimately the sequence $\hat{\pmb{z}}^4$ was selected.

In brief, by comparing the distribution of the same position base types of the sequence with that of the $K$-mers sequence, MRS Can find out the sequences with fewer base insertions, deletions, and substitutions, to find the sequences with high relative reliability. The pseudo-algorithm table of MRS is given as shown in Algorithm 1.

\begin{figure}[htbp]
  \centering
    \includegraphics[width=0.45\textwidth]{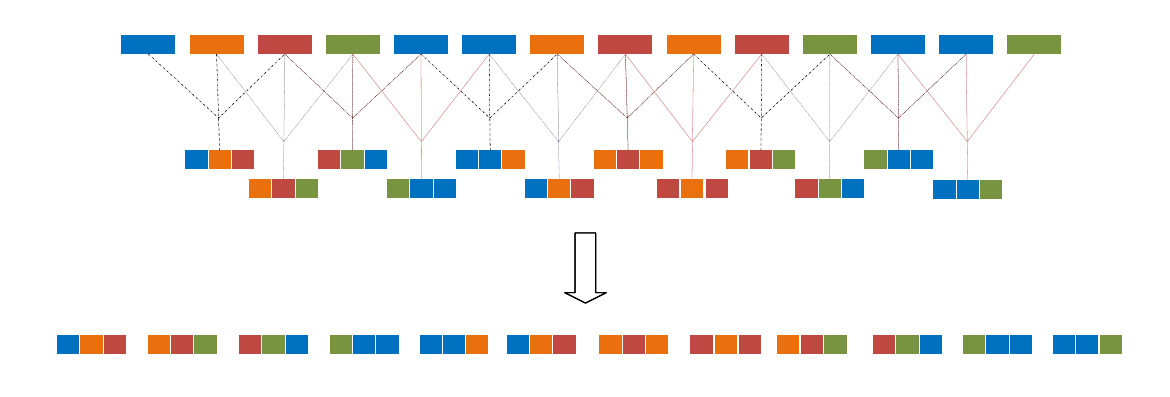}
    \caption{An example of the $K$-mers process with $K=3$.}
\label{fig.kmer}
\end{figure}

\subsection{Decoder}

The decoder consists of two steps: sequence analysis and image reconstruction. Firstly, the received multiple chains are subjected to convolutional operations with varying step sizes to analyze sequence information at multiple levels. The results are then concatenated and used for subsequent decoding. Subsequently,  convolutional layers are employed to reconstruct semantic segments of the image.
The decoding process can be delineated as 
\begin{align}\label{eq.chl3}
  \hat{\pmb{x}}=g_{\phi}(\{\hat{\pmb{z}}^{b_1},\hat{\pmb{z}}^{b_2},..., \hat{\pmb{z}}^{b_w}\}),
\end{align}
where $g_{\phi}:\mathbb{Z}^{w \times k} \xrightarrow{} \mathbb{R}^{n}$ denotes the decoding function and $\phi$ represents the parameter set of the decoder.\\

\subsection{Potential scenarios.}

The semantic extraction module of the SemAI-DNA storage system can be embedded in edge computing devices to directly process and encode data collected from IoT sensors. This reduces the need to transfer large amounts of raw data, thereby increasing the efficiency of data storage. Utilizing existing network infrastructure such as Wi-Fi and 5G, the encoded data can be efficiently transferred to DNA storage systems. During decoding, SemAI-DNA's multi-read filter model leverages the multi-copy properties of DNA molecules to enhance system fault tolerance and ensure the security and integrity of data transmission.\\

\section{END-TO-END TRAINING}
\label{sec.design}

In this section, we introduce the end-to-end implementation process of the SemAI-DNA.

The training framework of our model is illustrated in Fig.~\ref{fig.train}, which consists of two parts. After completing the training of the ISE, the generated semantic image dataset is used to perform the encoder-decoder training.

\subsection{ISE training}

SAM is a pre-trained,  large-scale AI model that does not require further training. It utilizes human-perceived semantic information and images as an experiential knowledge base to conduct foundational training for ASI. During ASI training, semantic segments are used as inputs to the network, with human-perceived information serving as a supervised training database. The model selects the choices most closely aligned with human perception from dispersed semantic information, generating images of semantic perception that are of greatest concern to humans.

\begin{figure}[htbp]
  \centering
    \includegraphics[width=0.5\textwidth]{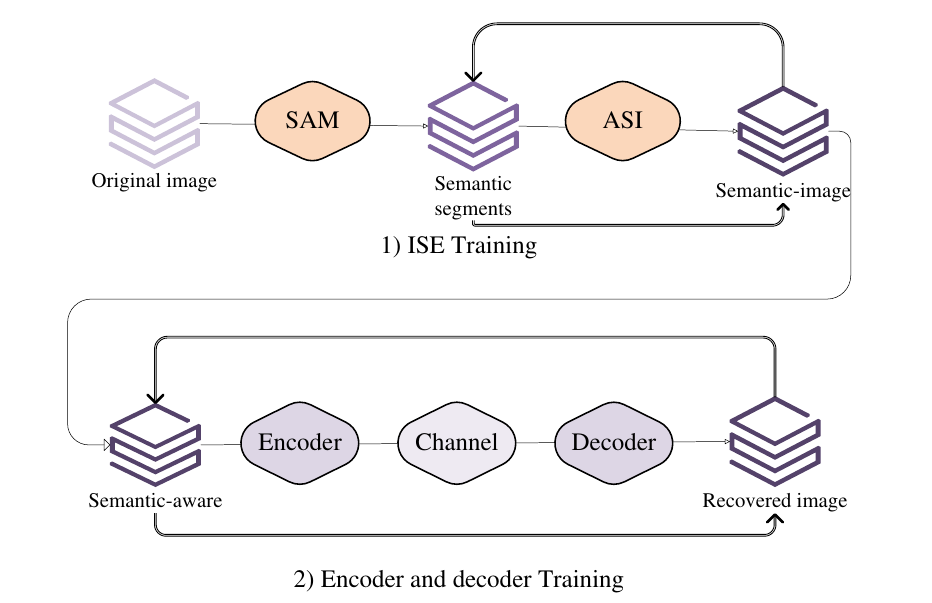}
    \caption{The training process.}
\label{fig.train}
\end{figure}

\begin{table}[tbp]
  \centering
  \caption{Network parameters}
  \begin{tabular}{l  l}
  \toprule
  Encoder                                       & Decoder                                             \\
    \midrule
  Conv2d (16,3,2), PReLU                             &   Conv1d (1,3,1), PReLU                             \\
  Conv2d (32,3,1), BN, PReLU      &   T-Conv2d (32,3,1), PReLU                          \\
  Conv2d (32,3,1), BN, PReLU                    &  T-Conv2d (32,3,1), BN, PReLU \\
  Conv2d (c,3,1)                                & T-Conv2d (16,4,2), BN, PReLU                        \\
  $round~(\text{Sigmoid}\times 3)$             &  T-Conv2d (3,4,2), Sigmoid                          \\

  \bottomrule
  \end{tabular}
  \label{tab.Structureparameters}
  \end{table}

\subsection{Encoder-Decoder training}

The functions that the encoder and decoder used in our methodology are listed in Tab.\ref{tab.Structureparameters}. Here, the tuple notation $(C, N, S)$ signifies a convolutional layer composed of $C$ filters, each with a kernel size $N$ and a stride $S$. The base pixel radio is represented as $R=k/n$. We can modulate $R$ by varying the value of $c$.
The DNA channel is constituted as non-trainable layers. In light of the present state of DNA synthesis techniques, synthesizing long-chain DNA is both labor-intensive and inefficient. Therefore, we partition the output sequence of the encoder into non-overlapping segments, each of length $s$, which are then recombined before being fed into the decoder.

The loss function comprises terms that ensure both image reconstruction quality and adherence to the biological constraints of the DNA chain. Generally, biological constraints encompass restrictions on GC content and polymer length. Typically, the GC content is maintained within the range of $40$-$60\%$, and the homopolymer run-length should not be excessively long, preferably less than 5~\cite{2013Characterizing,2017DNA}. 
The loss function is expressed as follows~\cite{2023Deep}:
\begin{align} 
\label{eq.Lossfunction}
    \mathcal{L}(\theta,\phi)=&\frac{1}{n} \sum_{i=1}^{n} (\pmb{x}_i,\hat{\pmb{x}}_i)^2 +
  \nonumber  \\
  &\alpha \times \frac{1}{m} \sum_{i=1}^{m} \left [ (\mathcal{T}_i,\mathcal{T}^*)^2+ (\mathcal{H}_i,\mathcal{H}^*)^2 \right],
\end{align} 
where $\alpha$ is an adjustable parameter. 
Here, the sequence $\pmb{z}$ is divided into $m$ overlapping segments. $\mathcal{T}_i$ and $\mathcal{H}_i$ respectively measure the base composition and base distribution of the $i${-th} segment of the DNA sequence, used to constrain the GC content and homopolymer length. $\mathcal{T}^*$ and $\mathcal{H}^*$ represent their expected values. The index of the base of $i$-th segment in the DNA sequence $\pmb{z}$ forms the set $\mathcal{M}_i$. 
\begin{subequations}\label{eq.th}
  \begin{align}
     \mathcal{T}_i &= \frac{1}{d}\sum_{t\in\mathcal{M}_i} \pmb{z}_t , \\
     \mathcal{H}_i &= \frac{1}{d-1}\sum_{t\in\mathcal{M}_i} ({\pmb{z}_t-\mathcal{T}_i})^2, \\
     \mathcal{T}^* &= 1.5,\\
     \mathcal{H}^* &= 1.25. 
  \end{align}
  \end{subequations}

\section{Numerical Results}
\label{sec.resultsbig}

\begin{table}[tbp]
  \centering
  \caption{Simulation parameters}
  \begin{tabular}{l l}
  \toprule
  Parameters   &  Values
   \\ \midrule
   Number of pixels $n=H\times W\times C$     & $64\times 64\times 3$ \\
  Base pixel radio $R=k/n$ (nts/pixel)  &  $[0.125,0.25,0.375,0.5]/3$ \\
  Base error rate $\gamma_{\mathrm{train}}$    &  $1\%$   \\
  The number of Multi-Reads $v$    &  $8$    \\
  The number of MRS $w$    &  $[2,4,8]$\\
  Length of segments $s$ & 256 \\
  Adjustable parameter in loss function $\alpha$  &   $1$   \\
Parameter in MSR  $K$  &   $7$   \\
  
  \bottomrule
  \end{tabular}
  \label{tab.parameter}
  \end{table}

In this section, we demonstrate the performance of our proposed SemAI-DNA model for DNA storage, with simulation settings discussed in Section~\ref{sec.simulationSetup} and numerical results given in Section\ref{sec.evaluationResults}
\subsection{Simulation Setup }
\label{sec.simulationSetup}

The system is implemented in Pytorch and the Adaptive Moment Estimation (Adam) optimizer is employed during the optimization process. The complete model undergoes training for 100,000 iterations, with an initial learning rate of $5\times 10^{-3}$, subsequently reduced to $5\times 10^{-4}$. The VOC2012 image dataset is employed for evaluating our scheme and resize the images to $64 \times 64$ size to speed up the training. Tab.~\ref{tab.parameter} lists the particulars of the simulation parameters. The effectiveness of this approach is gauged by evaluating the PSNR and SSIM.

"SemAI-DNA[w/o MRS]" represents the utilization of the SemAI-DNA without the inclusion of the MRS module. In this configuration, the functionality of the MRS module is replaced by randomly selecting a specific number of sequences for decoding. "SemAI-DNA[w/ MRS]" or "SemAI-DNA" indicates the adoption of the complete SemAI-DNA model, including the MRS module.

\subsection{Evaluation Results}
\label{sec.evaluationResults}

Fig.~\ref{fig.loss_epoch} illustrates the variation of loss values with increasing epochs. It can be observed that, under the same values of $\gamma_{\mathrm{train}}$ and $R$, the convergence results of the proposed SemAI-DNA scheme outperform the traditional DJSCC-DNA scheme~\cite{2023Deep}. The DJSCC-DNA used the structure is shown in Fig.~\ref{fig.model0}, with other settings consistent with the SemAI-DNA.

\begin{figure}[htbp]
  \centering
    \includegraphics[width=0.5\textwidth]{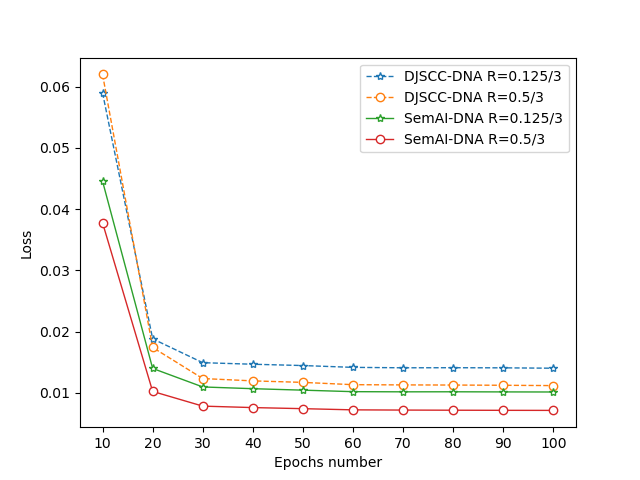}
    \caption{The loss performance over different number of epochs.}
\label{fig.loss_epoch}
\end{figure}

\begin{figure}[htbp]
  \centering
    \includegraphics[width=0.5\textwidth]{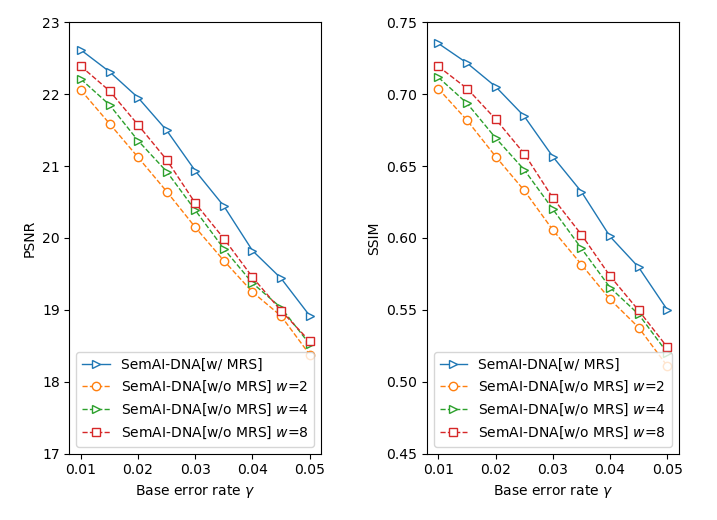}
    \caption{Performance of MRS in $R=0.25/3~(\mathrm{nt/pixel})$.}
\label{fig.compara}
\end{figure}

The simulation analysis is conducted on the performance of the proposed MRS. As shown in Fig.~\ref{fig.compara}, $w$ denotes the number of sequences used for decoding. When the channel pollution level is the same, with the increase of $w$ under the "SemAI-DNA[w/o MRS]" method, the quality of image restoration also improves. This is because the decoding end receives more information that can be used for decoding. Although this information is contaminated to varying degrees, the advantage of quantity can reduce the interference of erroneous information to some extent. It can be observed that as the channel pollution level increases, which is caused by higher error rates of sequences, it affects the decoding process and leads to a decline in the quality of image restoration. At the same time, the advantage of decoding multiple sequences becomes less apparent. This is because, under high error rates, the quantity advantage of sequences can easily burden the decoding process. The presence of frequent errors prompts the convolutional network to lower the requirements for image restoration in order to learn more valid information. As shown in the graph, when the "SemAI-DNA[w/ MRS]" method is employed with a setting of $w=4$, both the PSNR and SSIM of the restored images demonstrate better performance. The configuration of multi-reads selection filters out sequences that may disturb the decoding process in advance at the decoding end, thereby improving decoding performance.

Subsequently, the performance of the SemAI-DNA is compared with the DJSCC-DNA in several aspects.

\begin{figure}[htbp]
  \centering
    \includegraphics[width=0.5\textwidth]{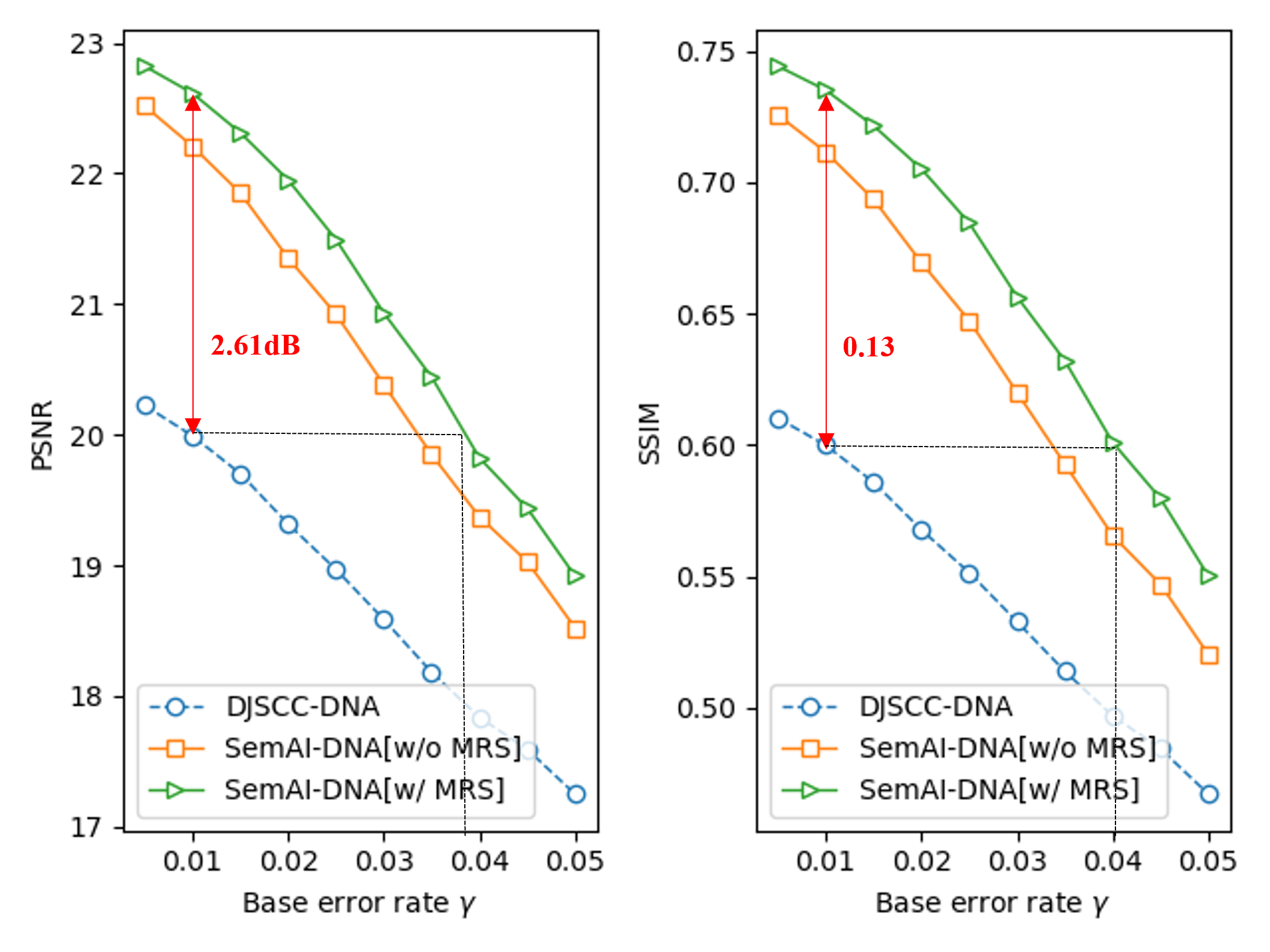}
    \caption{Performance of  SemAI-DNA  and DJSCC-DNA in $R=0.25/3~(\mathrm{nt/pixel})$.}
\label{fig.DNA2}
\end{figure}

As shown in Fig.~\ref{fig.DNA2}, the performance metrics of different schemes under different $\gamma$ are tested.
Analyzed from the perspective of vertical comparison, under the same $\gamma$, the "SemAI-DNA[w/ MRS]" and "SemAI-DNA[w/o MRS]" scheme are both much higher than the DJSCC-DNA scheme in terms of PSNR and SSIM of the image, that is, SemAI-DNA conveys the semantic information more accurately.
Analyzed from the perspective of side-by-side comparison, the higher $\gamma$ indicates the worse channel conditions for DNA storage, which acts as a hindrance to image restoration in both schemes. On the same PSNR or SSIM, the SemAI-DNA[w/ MRS] corresponds to a higher $\gamma$, which indicates that SemAI-DNA is able to resist worse storage channel conditions when conveying the same amount of information.
Overall, SemAI-DNA exhibits good image restoration quality. In particular, when the error rate is $0.01$, the PSNR is $2.61 ~\text{dB}$ higher and the SSIM is $0.13$ higher than that of the original algorithm.

Furthermore, to ensure biological constraints on the DNA chain, SemAI-DNA employs the same loss function as that used in DJSCC-DNA. We assessed the biological characteristics of the generated DNA chains, including homopolymer run-length and base compositions. As depicted in Fig.~\ref{fig.DNA4}(a), the x-axis represents the values of homopolymer run-length $i$, while the y-axis indicates the average proportion of homopolymers with length $i$ in the DNA chain. Both schemes predominantly restrict homopolymer run-length to $1$ and $2$, with proportions of homopolymer run-length exceeding length $4$ being less than $0.001$. Illustrated in Fig.~\ref{fig.DNA4}(b), the x-axis represents the base types, and the y-axis denotes the GC content in the DNA chain. It is observed that the GC content in both approaches is below $60\%$. SemAI-DNA not only improves the image storage performance but also maintains the biologocal constraints on the DNA chain, thereby ensuring its biological stability.

\begin{figure}[htbp]
  \centering
    \includegraphics[width=0.5\textwidth]{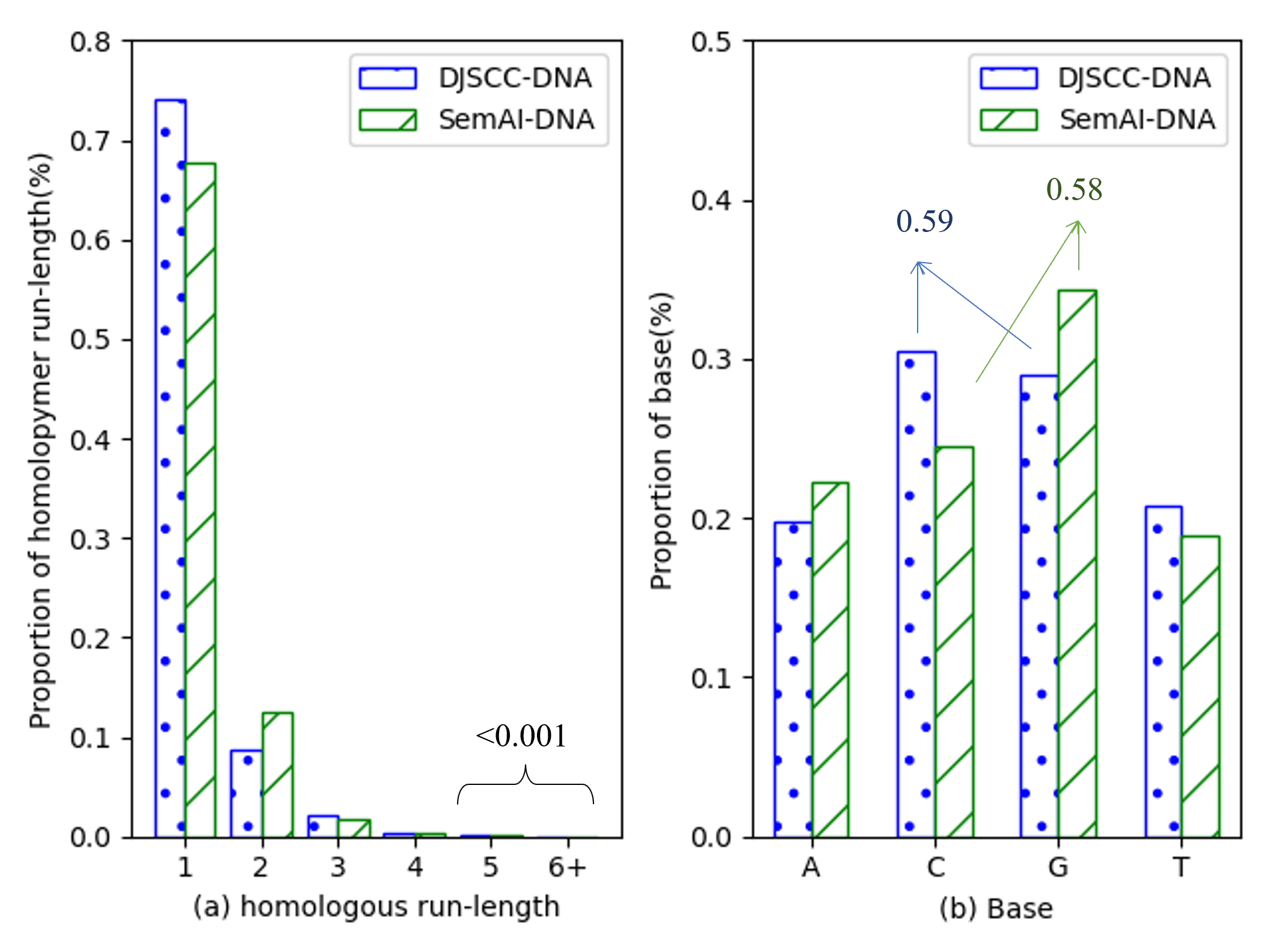}
    \caption{Biological characteristics of DNA chain of  SemAI-DNA and DJSCC-DNA in $R=0.25/3 ~(\mathrm{nt/pixel})$.}
\label{fig.DNA4}
\end{figure}

Finally, we simulated and compared the performance of several DNA storage systems. For the non-semantic DNA storage paradigm, we compared the DJSCC-DNA and VAEU-QC schemes. The VAEU-QC is a traditional coding strategy that uses the VAEU method~\cite{Generative} to achieve latent space representation of images and employs the quaternary coding~\cite{2019Author} scheme to store information in DNA sequences. Due to its higher requirements for channel coding and processing, we simulated its performance using a noise-free channel.
Additionally, to demonstrate the performance of the MRS module further, we simulated the ``SemAI-DNA[w/o MRS]", as we had done before.
As shown in Fig.~\ref{fig.rate_psnr}, SemAI-DNA[w/ MRS] exhibits notably higher PSNR and SSIM across various $R$. Furthermore, the MRS module also enhances performance to varying degrees.

\begin{figure}[htp]
  \centering
    \includegraphics[width=0.5\textwidth]{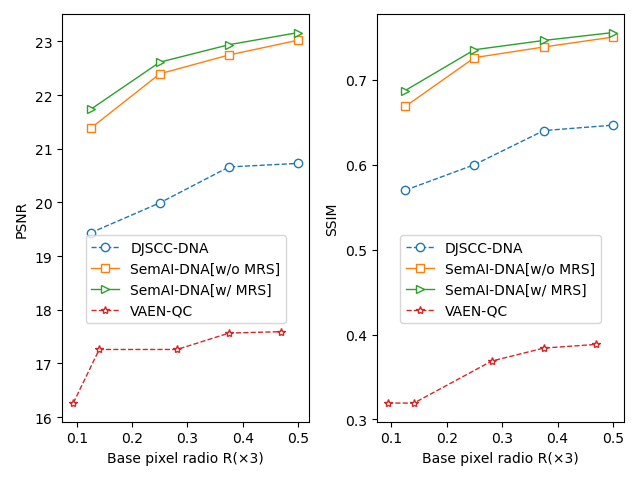}
    \caption{Performance comparison.}
\label{fig.rate_psnr}
\end{figure}

Furthermore, Fig.~\ref{fig.DNA3} presents the effect images at $R=0.5/3$. The images are first cropped to a size of $256 \times 256$, then divided into non-overlapping blocks for storage simulation and stitched restoration. The first row of the figure shows the original image. The second row displays semantic segments extracted by the ISE model, highlighting various image components. The third, fourth, and fifth rows present restored images from SemAI-DNA, DJSCC-DNA, and VAEU-QC, respectively, with corresponding PSNR and SSIM values.  As we can see SemAI-DNA has a better image storage effect and the borders generated by segmentation and stitching of image blocks are weaker than DJSCC-DNA and VAEU-QC. 
To summarize, the proposed SemAI-DNA framework achieves more effective storage of crucial semantic information under equivalent bandwidth constraints. This characteristic aligns more closely with actual storage requirements, providing a clearer and more accurate depiction of important image features while optimizing the use of available bandwidth.

\begin{figure}[htbp]
  \centering
    \includegraphics[width=0.5\textwidth]{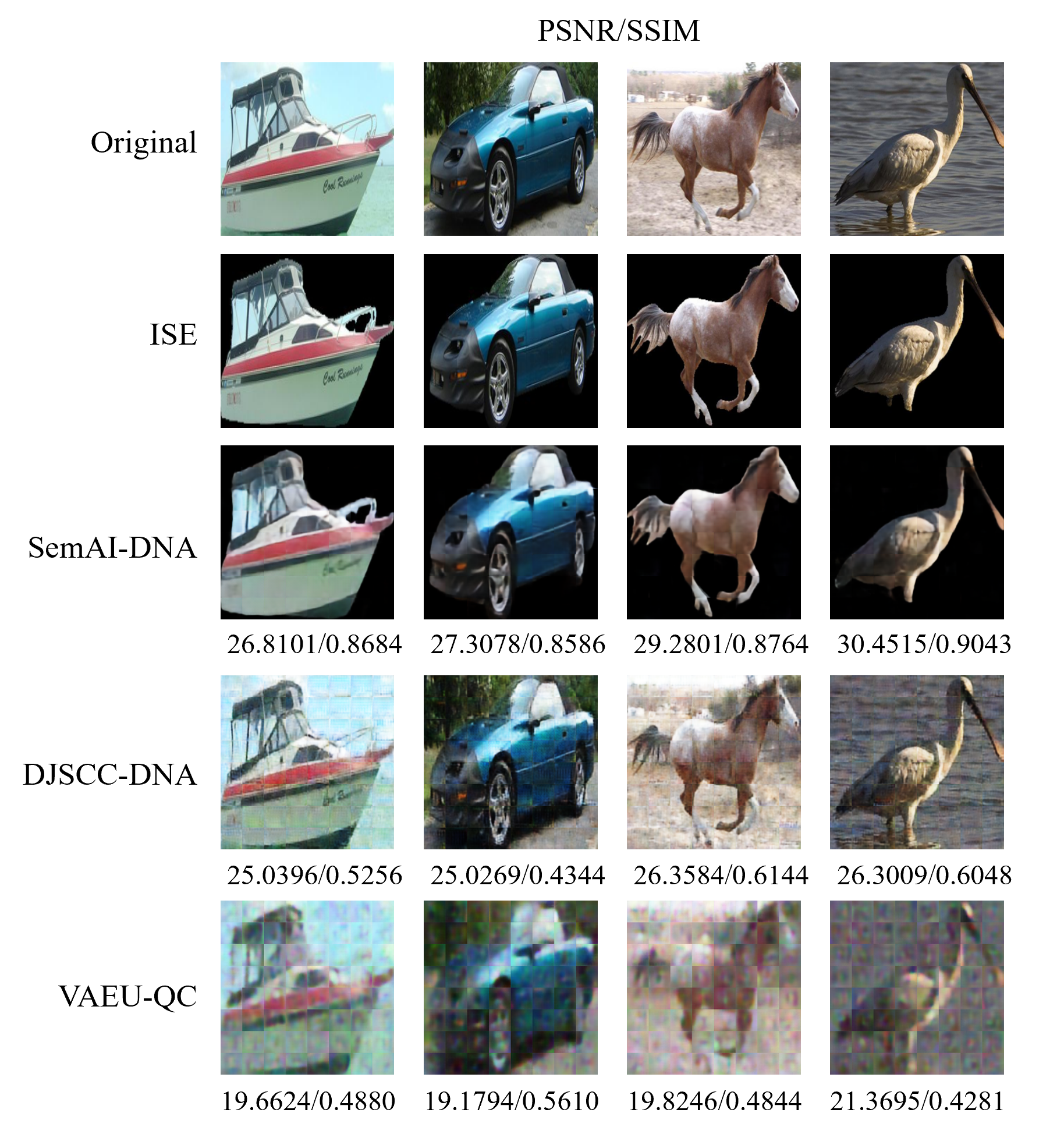}
    \caption{Examples of reconstructions for the SemAI-DNA.}
\label{fig.DNA3}
\end{figure}

\subsection{The complexity analysis}

The subsection provides a brief analysis of SemAI-DNA's complexity, comparing it with DJSCC-DNA. These simulations are performed on a server equipped with an Intel Xeon CPU, $128~\text{GB}$ of memory, and an RTX3080 GPU with $10 \text{GB}$ of video memory.

Tab.~\ref{tab.modelparams} and Tab.~\ref{tab.inference} compare the model complexity of SemAI-DNA with that of DJSCC-DNA, including the number of parameters, model storage size, and the time required for model inference on a single image. The difference between SemAI-DNA and DJSCC-DNA lies primarily in the sequence analysis portion of the decoder. Consequently, the parameters and storage space of this module are slightly higher compared to the latter. SemAI-DNA employs an ISE module based on a large-scale AI, resulting in a significantly larger number of parameters and required memory. Its advantage lies in achieving high performance with minimal training time costs by leveraging a large-scale AI. The average duration for semantic-aware image analysis using the ISE module inSemAI-DNA is $0.2698~\text{s}$. If an encoder-decoder module without MRS is used for image storage, the total duration required is $0.2806~\text{s}$, which is $0.1163~\text{s}$ faster than SemAI-DNA's $0.3969~\text{s}$. Although the number of model parameters between the encoder-decoder modules of SemAI-DNA and DJSCC-DNA are comparable, 
the inference is faster due to less redundancy of image data in SemAI-DNA.
When using an encoder-decoder module with MRS, the total duration is $0.2824~\text{s}$ seconds slower than the DJSCC-DNA scheme due to the longer time required for sequence comparison with MRS. However, the image quality with the MRS module is higher compared to without, as evidenced in Fig.~\ref{fig.DNA2} and Fig.~\ref{fig.rate_psnr}.

In summary, while SemAI-DNA exhibits high model complexity, its inference speed is comparable to traditional schemes, and it offers higher image storage performance.

\begin{table}[!ht]
  \centering
  \caption{Model analysis}
  \begin{tabular}{l c c c}
    \toprule
      Approach & Module & Total params & Total memory \\ \midrule
      SemAI-DNA& ISE & 637,037k & 612~MB \\ 
      SemAI-DNA& Encoder-Decoder & 114,401 & 0.57~MB \\ 
      DJSCC-DNA & Encoder-Decoder & 100,548 & 0.56~MB \\ 

       \bottomrule
  \end{tabular}
\label{tab.modelparams}
\end{table}

\begin{table}[!ht]
  \centering
  \caption{Model inference time (s)}
  \begin{tabular}{l c c c}
    \toprule
      Approach & ISE& Encoder-Decoder& Encoder-MRS-Decoder \\ \midrule

       SemAI-DNA& 0.2698 & 0.0108 & 0.4095 \\ 
       DJSCC-DNA & - & 0.3969 & - \\ 
       \bottomrule
  \end{tabular}
  \label{tab.inference}
\end{table}

\section{CONCLUSION AND FUTURE TRENDS}

This paper introduces a new DNA storage paradigm, namely SemAI-DNA for the upcoming IoT era.  A pair of key modifications have been proposed: embedding a semantic extraction module for the detailed encoding of semantic information and implementing the MRS model to enhance system fault tolerance. Numerical results have demonstrated its efficacy, showing a significant improvement over baseline models with a 2.61 dB gain in PSNR and a 0.13 increase in SSIM. 

SemAI-DNA presents an innovative and reliable solution to the challenges of data storage, particularly in the context of managing large-scale data. By leveraging advanced AI technology, this framework addresses the limitations of traditional storage methods in handling substantial volumes of data. In the context of Internet of Things (IoT) applications, SemAI-DNA holds potential for effectively managing and storing data generated by numerous devices. Future research should focus on expanding the capabilities of SemAI-DNA, including exploring improvements in data transmission and enhancing data security through DNA storage techniques. Additionally, while SemAI-DNA currently demonstrates efficacy in image storage, there is potential for extending its application to other data types, such as text and video, thereby increasing its utility in data storage for IoT devices.

\bibliographystyle{ieeetr}	
\bibliography{jscc-dna-bib}

\end{document}